\documentclass[conference]{IEEEtran}
\IEEEoverridecommandlockouts
\usepackage{cite}
\usepackage{amsmath,amssymb,amsfonts}
\usepackage[colorlinks,urlcolor=blue,linkcolor=blue,citecolor=blue]{hyperref}
\usepackage{graphicx}
\usepackage{textcomp}
\usepackage{xcolor}
\usepackage[numbers,sort&compress]{natbib}
\usepackage{color,array}
\usepackage{caption}
\usepackage{mathtools} 
\usepackage{soul}
\usepackage{fancyhdr}
\usepackage{subcaption}
\usepackage{multirow}
\usepackage{multicol}
\usepackage{enumitem}
\usepackage[noabbrev]{cleveref}
\usepackage{algorithm}
\usepackage{algpseudocode}

\fancypagestyle{morph-repo}{\fancyhf{} \fancyfoot[L]{**\url{https://github.com/lanl/MORPH}}}

\fancypagestyle{header_ipdps}{
  \fancyhf{}

  \fancyhead[C]{\footnotesize Accepted in 2025 IEEE International Parallel and Distributed Processing Symposium Workshops (IPDPSW)}
}

\begin{document}

\title{PDE foundation model-accelerated inverse estimation of system parameters in inertial confinement fusion\\
\thanks{Research presented in this article was supported by the Laboratory Directed Research and Development program of Los Alamos National Laboratory under project number 20250637DI. This research used resources provided by the Los Alamos National Laboratory Institutional Computing Program, which is supported by the U.S. Department of Energy National Nuclear Security Administration under Contract No. 89233218CNA000001.}
}


\author{
\IEEEauthorblockN{
Mahindra Rautela\IEEEauthorrefmark{1},
Alexander Scheinker\IEEEauthorrefmark{1},
Bradley Love \IEEEauthorrefmark{2},
Diane Oyen\IEEEauthorrefmark{2},
Nathan DeBardeleben\IEEEauthorrefmark{3}, \\
Earl Lawrence\IEEEauthorrefmark{2},
Ayan Biswas\IEEEauthorrefmark{2}
}
\IEEEauthorblockA{\IEEEauthorrefmark{1}\textit{AOT-IC Group, Los Alamos National Laboratory}, Los Alamos, NM, USA}
\IEEEauthorblockA{\IEEEauthorrefmark{2}\textit{CAI Division, Los Alamos National Laboratory}, Los Alamos, NM, USA}
\IEEEauthorblockA{\IEEEauthorrefmark{3}\textit{HPC Division, Los Alamos National Laboratory}, Los Alamos, NM, USA}
\IEEEauthorblockA{Emails: \{mrautela, ascheink, love, doyen, ndebard, earl, ayan\}@lanl.gov}
}

\maketitle
\thispagestyle{header_ipdps}

\begin{abstract}
PDE foundation models are typically pretrained on large, diverse corpora of PDE datasets and can be adapted to new settings with limited task-specific data. However, most downstream evaluations focus on forward problems, such as autoregressive rollout prediction. In this work, we study an inverse problem in inertial confinement fusion (ICF): estimating system parameters (inputs) from multi-modal, snapshot-style observations (outputs). Using the open \emph{JAG benchmark}, which provides hyperspectral X-ray images and scalar observables per simulation, we finetune the PDE foundation model and train a lightweight task-specific head to jointly reconstruct hyperspectral images and regress system parameters. The fine-tuned model achieves accurate hyperspectral reconstruction (test MSE $1.2\times10^{-3}$) and strong parameter-estimation performance (up to $R^2=0.995$). Data-scaling experiments (5\%--100\% of the training set) show consistent improvements in both reconstruction and regression losses as the amount of training data increases, with the largest marginal gains in the low-data regime. Finally, finetuning from pretrained MORPH weights outperforms training the same architecture from scratch, demonstrating that foundation-model initialization improves sample efficiency for data-limited inverse problems in ICF.

\end{abstract}

\begin{IEEEkeywords}
Scientific Machine Learning, PDE Foundation Model, Plasma Physics, Inertial Confinement Fusion, Inverse problems
\end{IEEEkeywords}

\section{Introduction}

Standalone PDE surrogate models have dominated scientific machine learning for a variety of physical applications. Some of the popular methods include physics-informed neural networks (PINNs) \citep{RAISSI2019686}; operator-learning methods such as DeepONet \citep{goswami2022deep} and neural operators including Fourier neural operators (FNOs) \citep{li2020fourier}, wavelet neural operators (WNOs) \citep{tripura2023wavelet}, Laplace neural operators \citep{cao2024laplace}, and physics-informed neural operators (PINOs) \citep{li2024physics}; latent-state and reduced-order evolution models \citep{oommen2022learning,maulik2021reduced,rautela2024conditional,kontolati2024learning,rautela2025time}; and continuous-depth dynamical models such as Koopman-based formulations with nonlinear forcing \citep{khodkar2021data} and neural ODEs (ODE-Nets) \citep{chen2018neural}. Despite their success, these models are typically developed and trained for a specific PDE, parameterized family, geometry, or discretization, and often require retraining (or substantial re-tuning) when the governing physics, resolution, or observational modality changes.

This motivates pretraining-based approaches that aim to amortize learning across PDE families and enable more data-efficient adaptation to new tasks and regimes. PDE foundation models are task-agnostic neural surrogates pretrained on large and diverse collections of PDE solution datasets to learn transferable representations, and then adapted via finetuning to specific downstream tasks \cite{subramanian2023towards}. Relative to training from scratch, this pretraining--finetuning paradigm can improve sample efficiency and generalization, especially when transferring to previously unseen PDE families, parameter regimes, or geometries \cite{hao2024dpot,herde2024poseidon}. MORPH extends this idea to multi-modal scientific data by supporting heterogeneous PDE datasets across 1D--3D, varying spatial resolutions, and multiple physical fields with mixed scalar and vector components within a unified representation \cite{rautela2025morph}.

\begin{figure*}
	\centering
	\includegraphics[width=1.0\textwidth]{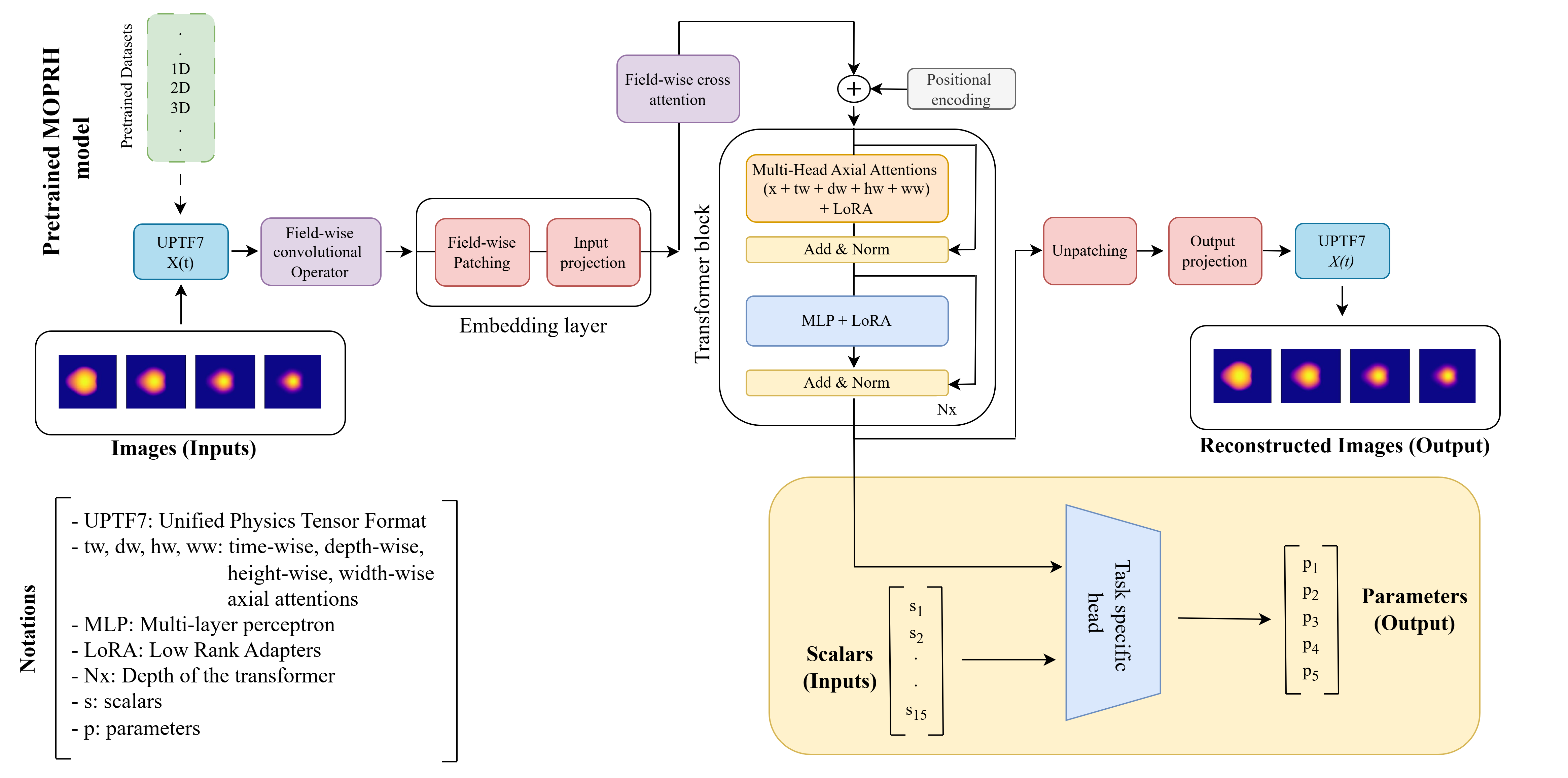}
	\caption{Schematic illustrating how the PDE foundation model (MORPH) latent representation is coupled with a lightweight task-specific head. Hyperspectral images are provided as input to the foundation model, which is fine-tuned for reconstruction. The transformer-block outputs are passed to a task-specific head (a dense neural network). In parallel, the task-specific head (TSH) also ingests 15 scalar observables/diagnostics and is trained to predict a 5D parameter output. The foundation model and TSH are trained jointly end-to-end, using separate loss functions with independent optimizers and learning-rate schedulers.}
	\label{fig:morph_ft}
	\vspace{2mm}
\end{figure*}

Inverse problems seek to infer latent causes such as material properties, boundary/initial conditions, or system/design parameters from indirect and often noisy observations generated by an underlying forward model \cite{vogel2002computational}. In contrast to forward prediction, the inverse mapping is frequently ill-posed: the solution may be non-unique, unstable to perturbations in the data, or may not exist without incorporating additional prior information or regularization. Consequently, practical inversion typically relies on constraints, explicit regularizers, and/or sensitivity-informed diagnostics to obtain stable and physically plausible estimates \cite{rautela2022inverse}.

While PDE foundation models have shown strong transfer and sample efficiency across forward modeling tasks, their use for inverse problems remains comparatively less explored. Most existing PDE foundation models are pretrained with autoregressive objectives on spatiotemporal datasets, and evaluation/finetuning commonly targets long-horizon rollouts \cite{mansingh2025towards}. Although inverse problems are not yet a primary focus in the PDE foundation-model literature, foundation models in other domains have been adapted to inverse tasks by attaching lightweight, task-specific heads (or decoders) on top of pretrained representations and finetuning only a small subset of parameters. This paradigm is well established in language \cite{brown2020language} and vision \cite{oquab2023dinov2}. More recently, it has become increasingly common in scientific foundation models, for example in chemistry and Earth-system modeling where pretrained backbones are repurposed for parameter inference, design, or reconstruction from partial observations \cite{wadell2025foundation,bodnar2025foundation}.

Recent works have explored a broad range of inverse problems in ICF. Humbird \emph{et al.} \cite{humbird2022transfer} use Bayesian optimization to select drive inputs that maximize neutron yield with only a small number of experiments. Wang \emph{et al.} \cite{wang2024multifidelity} extend this idea with iterative multi-fidelity Bayesian optimization, using Gaussian-process surrogates to combine low- and high-fidelity evaluations while dynamically allocating simulation budget. For post-shot analysis, Gaffney \emph{et al.} \cite{gaffney2024data} employ an MCMC-based hierarchical Bayesian calibration framework to infer latent simulation inputs (and their shot-to-shot variability) that make radiation-hydrodynamics simulations consistent with sparse NIF diagnostic observations. On the diagnostic inversion side, Serino \emph{et al.} \cite{serino2025physics} estimate physical/initial-condition parameters of a shock-driven instability experiment from noisy time-resolved radiographs using a feature-extraction-based, two-network learning pipeline. More recently, Gutierrez \emph{et al.} \cite{gutierrez2025lpg} study inverse design of laser pulse shapes by mapping desired implosion outcomes (and target parameters) to feasible drive profiles using diffusion and autoregressive generative models. In parallel, language-based foundation models are beginning to be applied to design-oriented ICF tasks, including spherical shell design and semi-empirical tuning to support experimental planning \cite{bahukutumbi2025fine,grosskopf2025ursa}.

In this work, we study an inverse problem estimating latent inertial confinement fusion (ICF) design parameters from multi-modal diagnostic observations and assess whether PDE foundation-model pretraining improves inversion accuracy and data efficiency relative to training from scratch. In the ICF setting considered here, the inverse task is to estimate the simulator input parameters from a multi-modal diagnostic signature consisting of hyperspectral images and scalar observables \cite{anirudh2020improved}. The key contributions of this work are:

\begin{enumerate}
    \item \emph{PDE foundation models for inverse problems.} We present (to our knowledge) one of the first demonstrations of adapting a PDE foundation model (MORPH) to an \emph{inverse} task, moving beyond the forward-rollout settings that dominate prior PDE foundation-model evaluations.

    \item \emph{ICF parameter estimation from multi-modal diagnostics.} We formulate and evaluate an ICF inverse problem on the JAG benchmark by finetuning MORPH to estimate simulator/design parameters from multi-modal observations comprising hyperspectral X-ray images and scalar diagnostics.

    \item \emph{Data-driven sensitivity analysis.} We introduce an interpretable sensitivity analysis (PCA compression of images + ridge regression over image PCs and scalars) to quantify parameter–observable dependence and identify weakly constrained (ill-posed) parameter directions.

    \item \emph{Systematic evaluation: scaling and scratch baselines.} We report data-scaling results (5\%--100\% training data) and a controlled training-from-scratch baseline using the \emph{same} architecture, isolating the benefit of foundation-model initialization and showing the largest gains in the low-data regime.
\end{enumerate}

\section{Methods}
\subsection{Dataset}
We use the open ICF dataset released in the LLNL \texttt{macc} repository, generated with a 1D semi-analytic ICF simulator that models capsule implosion dynamics and produces a multi-modal suite of synthetic diagnostics \cite{JAG_LLNL}. This dataset is commonly referred to as the JAG benchmark and has been used in Refs.~\cite{anirudh2020improved,anirudh2019exploring,jacobs2019parallelizing,shukla2025design}. Each simulation is specified by a five-dimensional input vector of design/physics parameters and yields two complementary observation modalities: (i) four spectrally resolved X-ray images (four energy bands) at $64\times64$ resolution and (ii) 15 scalar diagnostic quantities (e.g., yield, ion temperature, pressure, and related observables). We denote the parameter vector by $x \in \mathbb{R}^{5}$, the imaging observations by $y_{\mathrm{img}} \in \mathbb{R}^{64\times64\times4}$, and the scalar diagnostics by $y_{\mathrm{sc}} \in \mathbb{R}^{15}$. The dataset contains 10{,}000 samples generated by the simulator, which defines a forward mapping
$f:\, x \mapsto y = (y_{\mathrm{img}}, y_{\mathrm{sc}})$.

\subsection{Model}
PDE foundation models are pretrained on large, heterogeneous corpora spanning multiple PDE benchmarks, and can be fine-tuned in data-limited regimes where they often outperform identical architectures trained from scratch \cite{marcato2025foundation}. However, scaling pretraining across diverse benchmarks introduces a central challenge: learning from multiple data modalities while handling variation in dimensionality, resolution, and field structure (including both scalar and vector-valued quantities) across PDE datasets. Naively padding all samples to a common shape is often infeasible at scale because simulation corpora can be extremely large (e.g., The Well is 15 TB) \cite{ohana2024well}. To address this, MORPH was designed to enable multi-modal training over larger and more diverse PDE corpora while keeping computation tractable \cite{rautela2025morph}.

Built on \emph{Unified Physics Tensor Format} (UPTF-7), \textsc{MORPH} is a modality-agnostic architecture that addresses data heterogeneity through three complementary mechanisms (Fig.~\ref{fig:morph_ft}). First, \emph{component-wise convolutions} operate over scalar and vector components to inject locality biases that improve sample efficiency. Second, \emph{inter-field multi-head cross-attention} explicitly models coupling among distinct physical fields, enabling selective information transfer and robust feature fusion even when fields evolve on mismatched spatial/temporal scales; it can also reduce cost by producing a single fused field representation. Finally, \textsc{MORPH} uses \emph{4D axial attention} that factorizes spatiotemporal self-attention across time and spatial axes, reducing complexity from $O((TDHW)^2)$ to $O(T^2 + D^2 + H^2 + W^2)$ while retaining expressivity. 

MORPH is pretrained with an autoregressive objective that predicts the next time step from previous states, enabling it to learn transferable spatiotemporal representations. Pretraining is performed on a heterogeneous corpus drawn from widely used PDE benchmarks \cite{takamoto2022pdebench, ohana2024well, herde2024poseidon}, and the resulting model is then fine-tuned on multiple downstream forward-modeling tasks. The pretraining suite comprises six spatiotemporal datasets spanning 1D–3D domains with diverse field and component structure: 1D-CFD (computational fluid dynamics), 2D-DR (diffusion–reaction), 2D-SW (shallow water), 2D-CFD-IC (incompressible CFD/Navier–Stokes), 3D-MHD (magnetohydrodynamics), and 3D-CFD (computational fluid dynamics). In contrast, the physics governing ICF fusion and its hyperspectral X-ray signatures is not represented in this pretraining suite \cite{gaffney2014thermodynamic}. We therefore consider hyperspectral reconstruction and parameter inference for ICF as an out-of-distribution transfer setting. The codebase (including the model implementation and trained weights) for MORPH is open-sourced and publicly available in the associated repository**.
\thispagestyle{morph-repo}

We leverage the pretrained MORPH backbone and attach a lightweight, task-specific head (TSH) to estimate system parameters from hyperspectral images and scalar diagnostics (Fig.~\ref{fig:morph_ft}). The hyperspectral image tensor is processed by MORPH under an image-reconstruction objective, producing reconstructed images at the output. Alongside the reconstructed output, we use the final latent representation at the end of the transformer blocks as the image embedding for parameter regression. The image embedding is passed through two \texttt{Conv1D}$\rightarrow$\texttt{GELU}$\rightarrow$\texttt{Dropout} blocks, followed by two \texttt{Linear}$\rightarrow$\texttt{GELU}$\rightarrow$\texttt{Dropout} blocks. In parallel, the 15-dimensional scalar diagnostics are encoded with a smaller MLP and concatenated with the resulting image features. The resulting fused representation is then fed to a final projection layer to predict the system parameters.

\subsection{Sensitivity analysis}
\begin{figure*}
	\centering
	\includegraphics[width=1.0\textwidth]{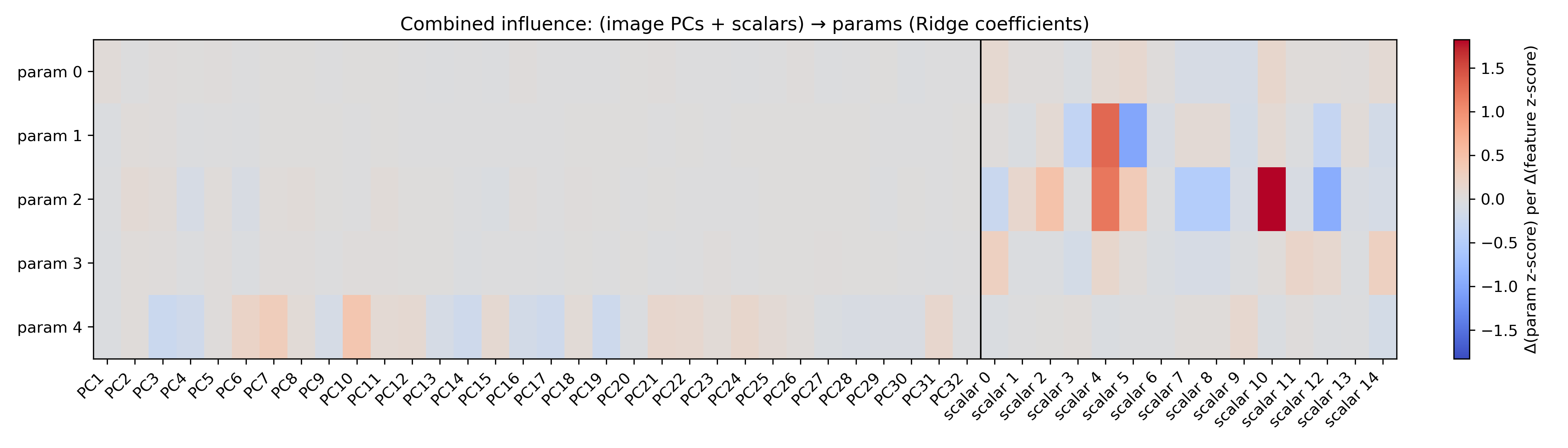}
	\caption{Ridge-regression sensitivity map for predicting the five parameters from multi-modal features. Columns show standardized image PCA scores (PC1--PC32) and scalar diagnostics (scalar0--scalar14), separated by the vertical line; colors denote signed coefficient magnitude. The strongest dependencies are scalar-driven for \texttt{param1} and \texttt{param2}, while \texttt{param0} and \texttt{param3} exhibit near-zero coefficients, indicating poor identifiability.}
	\label{fig:sensitivity_heatmap}
	\vspace{2mm}
\end{figure*}
Before implementing the fine-tuning pipeline, we perform a sensitivity analysis to quantify how strongly the outputs depend on the input parameters \cite{balasubramaniam1998inversion}. This is a useful preprocessing step for inverse problems because it provides an empirical indicator of ill-posedness: weak or non-unique dependence of the observables on certain parameters suggests poor identifiability and an inherently under-constrained inversion. By highlighting which observables carry information about which inputs, sensitivity analysis helps diagnose ill-posed parameter directions and guides model design and target selection, even without detailed expert knowledge of the simulator \cite{rautela2022inverse}.

We study a data-driven linear sensitivity analysis for an inverse problem with multi-modal observations.
Each sample comprises an image-like tensor $I \in \mathbb{R}^{64 \times 64 \times 4}$ together with a vector of scalar descriptors $s \in \mathbb{R}^{15}$, and the objective is to infer a low-dimensional parameter vector $y \in \mathbb{R}^{5}$.
Because the image modality is high-dimensional, we vectorize $I$ into $x \in \mathbb{R}^{64 \cdot 64 \cdot 4}$ and apply feature-wise standardization. This normalization removes scale effects, improves numerical conditioning, and makes regression coefficients comparable across heterogeneous modalities.

To obtain a compact and interpretable representation of the image channel, we apply principal component analysis (PCA) to the standardized image vectors and retain the leading $K$ components, yielding latent image coordinates $z \in \mathbb{R}^{K}$. We then employ ridge regression to construct stable linear relationships between variables of interest.
Ridge regularization (equivalently, Tikhonov regularization) mitigates ill-conditioning and multicollinearity, which commonly arise in high-dimensional inverse problems, and provides a well-posed surrogate for local, linearized sensitivity in the reduced space \cite{om2001ridge}.

To quantify the joint contribution of both modalities to parameter inference, we concatenate the reduced image features and scalar descriptors into a single feature vector $h = [z; s] \in \mathbb{R}^{K+15}$ and fit a ridge model mapping $h$ to standardized parameters.
The resulting coefficient matrix enables transparent attribution: weights associated with PCA coordinates summarize image-driven dependence, whereas weights associated with the scalar entries capture the influence of auxiliary measurements.
\section{Results}
\begin{figure*}[t]
    \centering

    \begin{subfigure}[b]{0.45\linewidth}
        \centering
        \includegraphics[width=\linewidth]{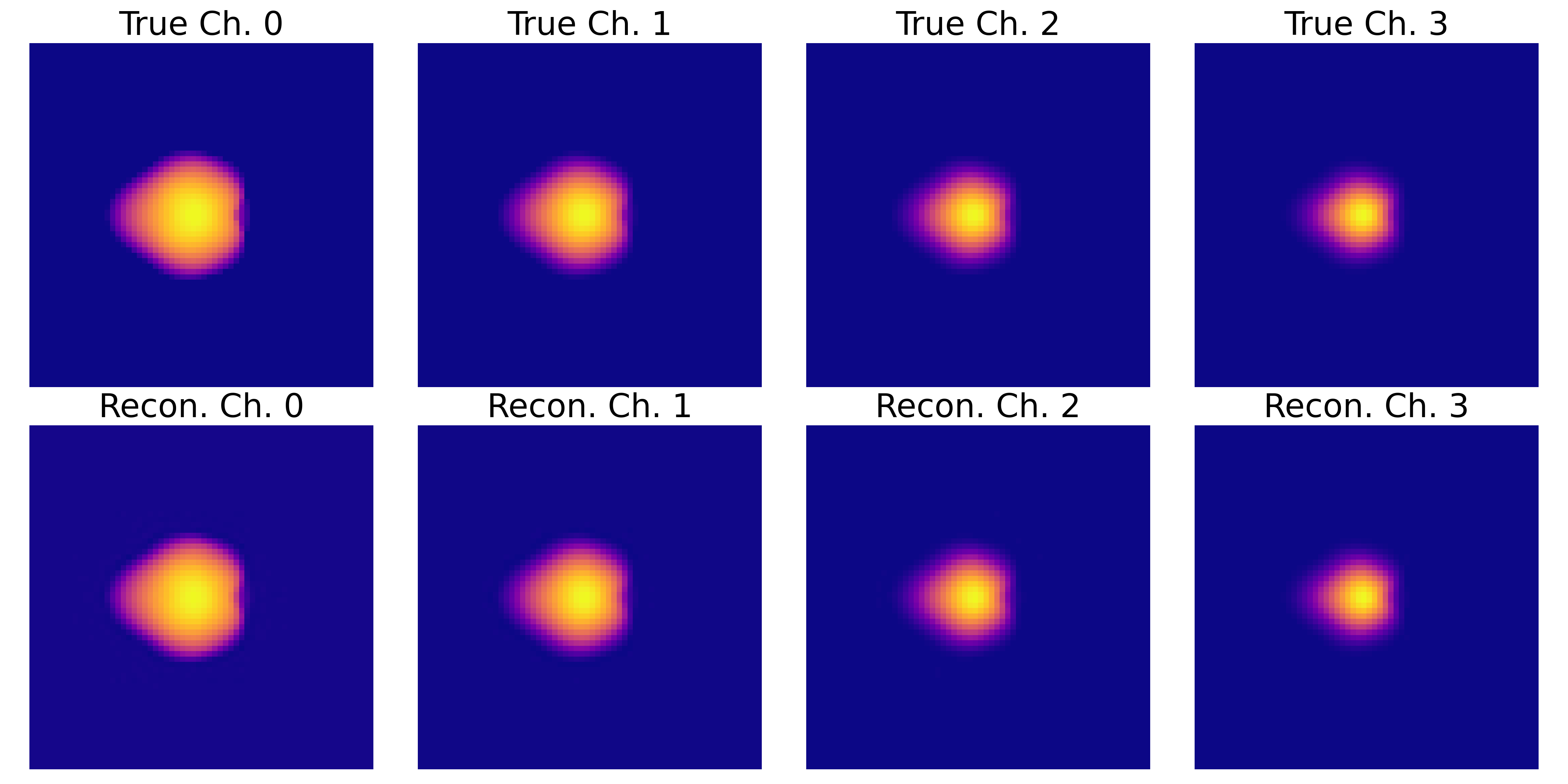}
        \caption{}
        \label{fig:recon_a}
    \end{subfigure}
    \begin{subfigure}[b]{0.45\linewidth}
        \centering
        \includegraphics[width=\linewidth]{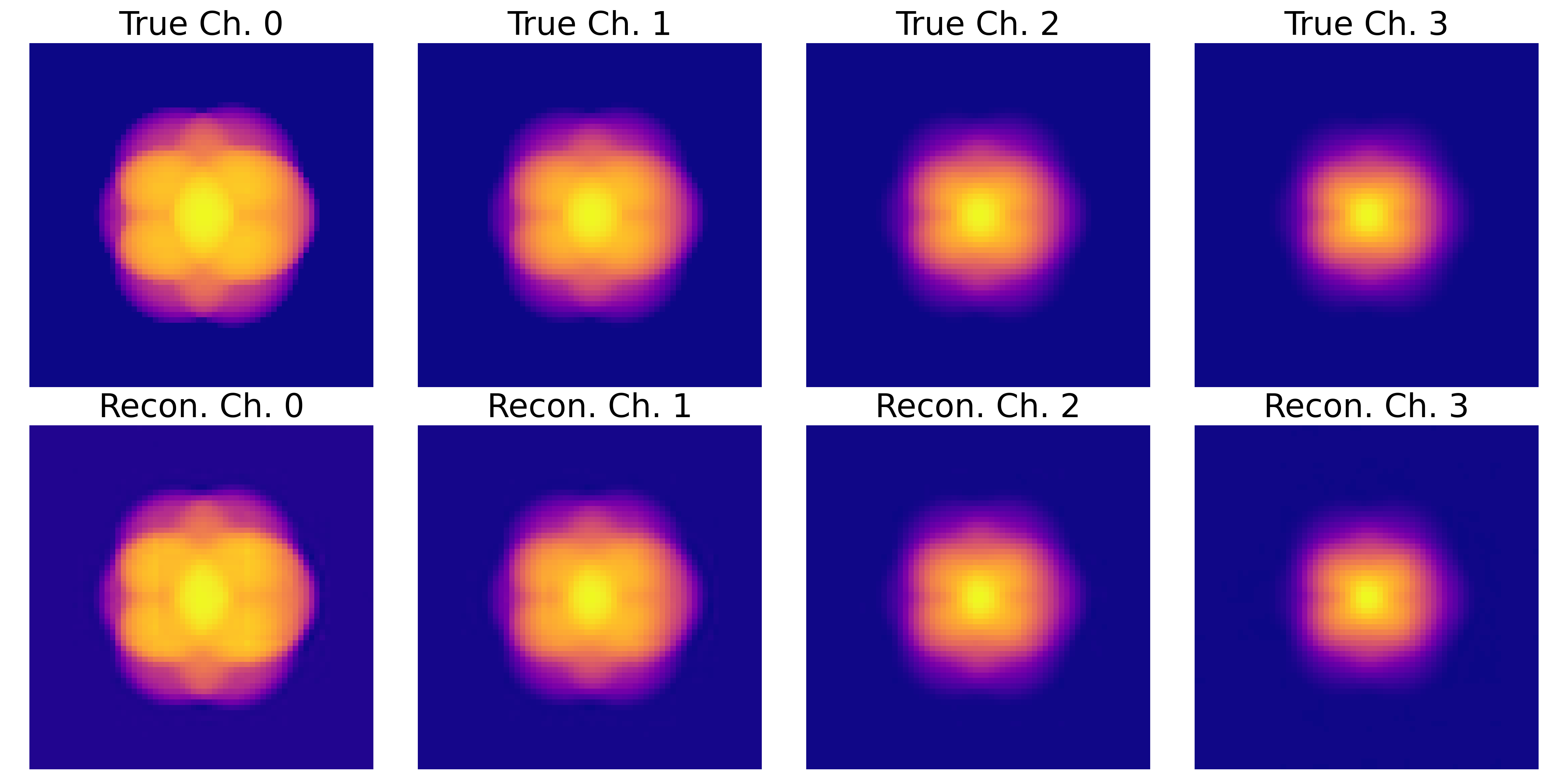}
        \caption{}
        \label{fig:recon_b}
    \end{subfigure}
    
    \begin{subfigure}[b]{0.45\linewidth}
        \centering
        \includegraphics[width=\linewidth]{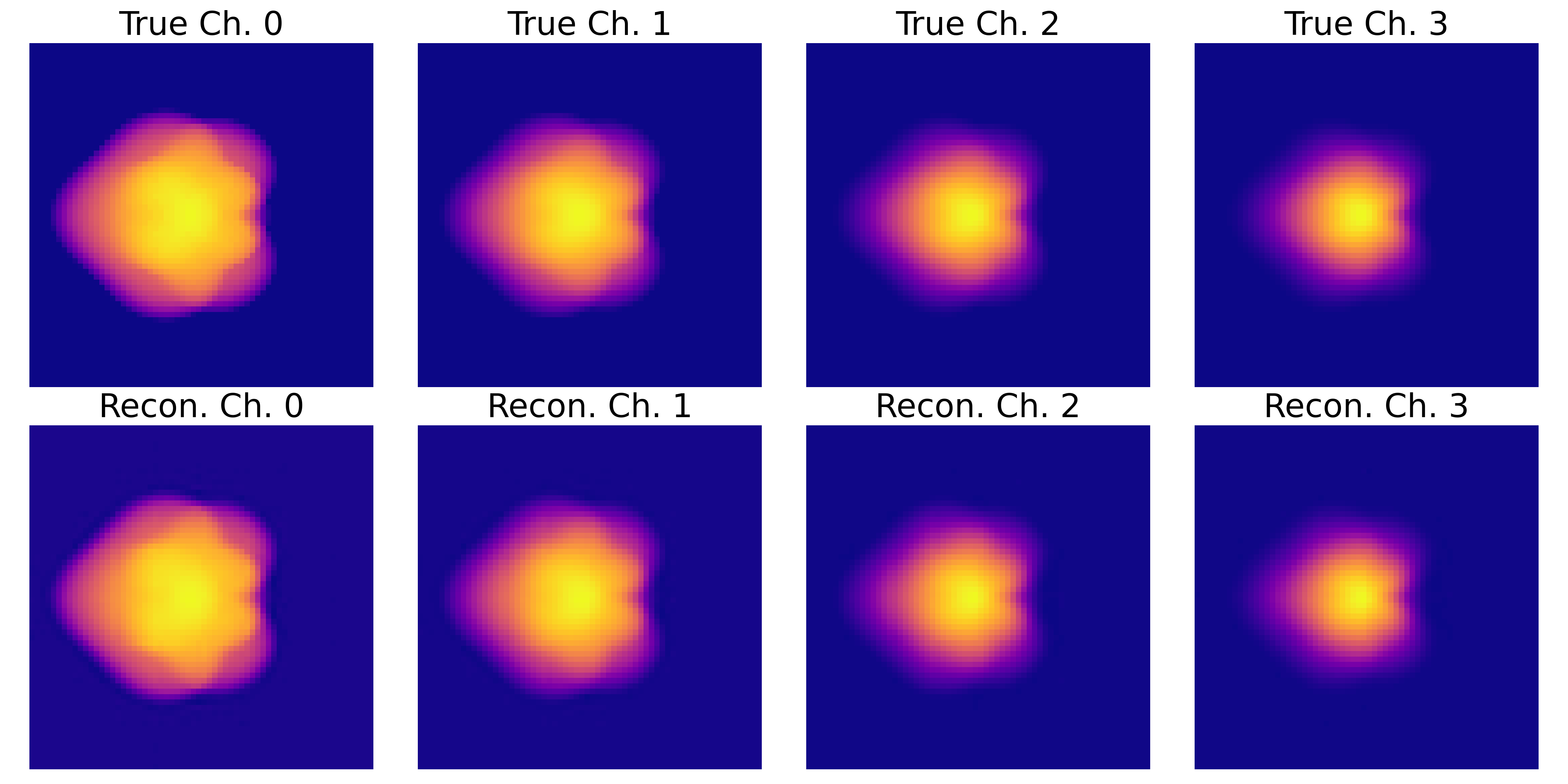}
        \caption{}
        \label{fig:recon_c}
    \end{subfigure}
    \begin{subfigure}[b]{0.45\linewidth}
        \centering
        \includegraphics[width=\linewidth]{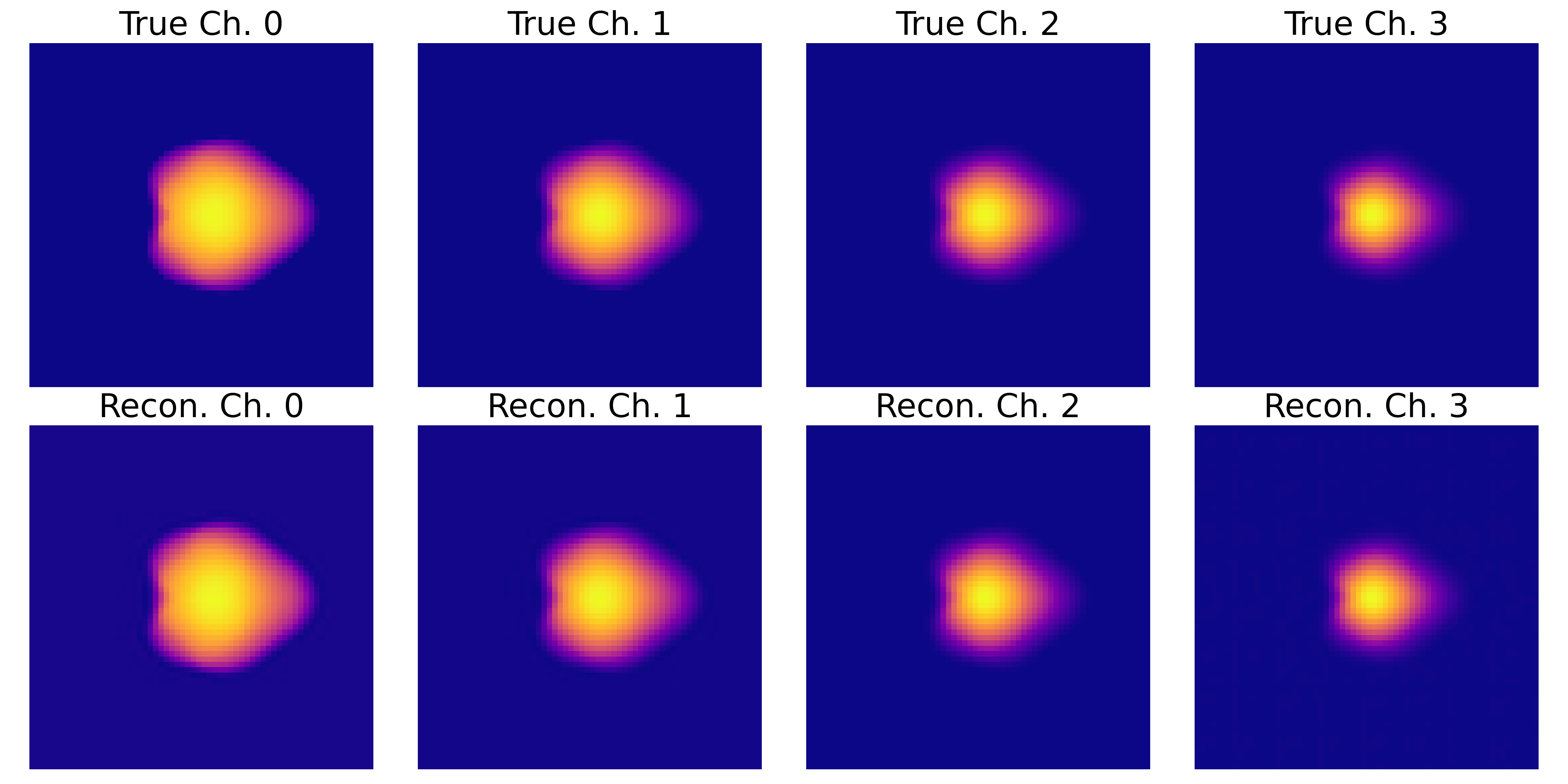}
        \caption{}
        \label{fig:recon_d}
    \end{subfigure}

    \caption{Reconstruction results on four held-out test samples (a-d). For each example, the top row shows the ground-truth four-channel hyperspectral image, and the bottom row shows the corresponding reconstructed (predicted) image. Across the full test set, the recorded MSE between true and reconstructed images is 0.0012.}
    \label{fig:recon}
\end{figure*}
\begin{figure*}
	\centering
	\includegraphics[width=0.95\textwidth,trim=0cm 0cm 0cm 0cm, clip]{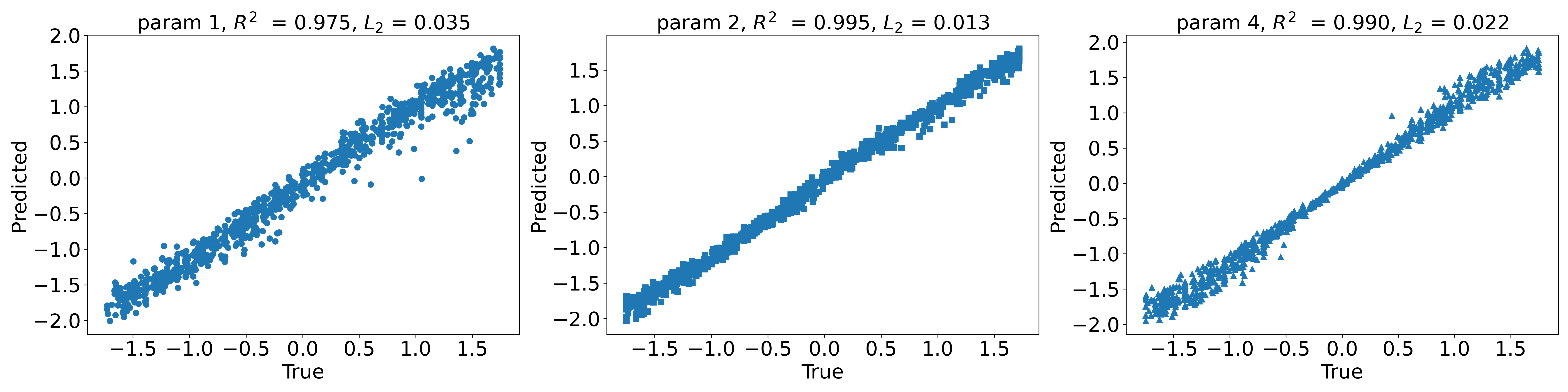}
	\caption{Parameter estimation results on the test set. Scatter plots compare ground-truth versus predicted values for each of the three predicted parameters. We achieve strong agreement for three parameters: Param1 attains $R^2=0.975$ with $L_2=0.035$, Param2 attains $R^2=0.995$ with $L_2=0.013$, and Param4 attains $R^2=0.990$ with $L_2=0.022$.}
	\label{fig:param_estimation}
	\vspace{2mm}
\end{figure*}

\subsection{Sensitivity results}

Fig.~\ref{fig:sensitivity_heatmap} summarizes the learned ridge coefficients for the combined model (image principal components + scalars) $\rightarrow$ parameters. Overall, parameters exhibit markedly different levels of sensitivity to the available observables. Parameters $\texttt{param1}$ and $\texttt{param2}$ show clear, structured dependence on a subset of scalar descriptors, with several coefficients attaining substantially larger magnitude than the background level. In contrast, the image PCA coordinates contribute weakly for these parameters, suggesting that the scalar channel carries most of the predictive signal in this linearized sensitivity analysis. 

The image modality contributes more visibly to $\texttt{param4}$, which displays non-negligible weights across multiple PCA components (albeit with smaller magnitude than the strongest scalar effects observed for $\texttt{param2}$). This pattern indicates that $\texttt{param4}$ is at least partially encoded in spatial/field structure captured by the dominant PCA modes, whereas the scalar descriptors play a comparatively minor role.

Most importantly, $\texttt{param0}$ and $\texttt{param3}$ exhibit near-zero coefficients across both feature blocks, indicating weak dependence on the measured observables under this linear surrogate. Consistent with this observation, these parameters achieve very low held-out $R^2$ compared to the remaining parameters, implying poor predictability and limited identifiability from the current observation set. In practice, this suggests that accurately recovering $\texttt{param0}$ and $\texttt{param3}$ will be difficult without additional information (e.g., more informative scalar diagnostics, richer image features, alternative measurement modalities, or stronger priors, larger training set), and motivates treating their inversion as intrinsically more ill-posed than the other parameters.

In prior studies \cite{anirudh2020improved,shukla2025design}, substantially larger training sets are available (on the order of $10^5$ samples), whereas the public repository used here provides only $\sim 10^4$ samples. Access to more data can improve the effective posedness of the inverse problem by better constraining the parameter--observation map, expanding coverage of the parameter space (and thus the diversity of observable responses), and reducing estimator variance. Motivated by our sensitivity analysis, we therefore focus on a reduced inversion task and predict only three of the five parameters, excluding the two parameters that exhibit consistently weak signal and poor prediction performance.

\subsection{Finetuning results}
Among the four open-source MORPH foundation-model variants: MORPH-Ti ($\sim$7M parameters), MORPH-S ($\sim$30M), MORPH-M ($\sim$126M), and MORPH-L ($\sim$480M), we have selected MORPH-S for this study. We fine-tune all parameters of MORPH-S and train the task-specific head (TSH), which has $\sim$1M parameters, for a total of 100 epochs on a single RTX-A6000 GPU. The 10K-sample dataset is randomly split into train/val/test with an 80\%/10\%/10\% ratio.

For both MORPH and TSH, we use the AdamW optimizer with weight decay 0.01, with learning rates of $10^{-4}$ for MORPH and $10^{-5}$ for TSH. We apply the same learning-rate schedule to both components: linear warmup for 5 epochs (minimum LR $10^{-7}$) followed by cosine decay. We use a batch size of 8 and optimize mean-squared error (MSE) losses for both the reconstruction and parameter-regression objectives. 

\begin{figure*}[t]
    \centering
    \captionsetup[subfigure]{justification=centering}
    \begin{subfigure}[b]{0.24\linewidth}
        \centering
        \includegraphics[width=\linewidth]{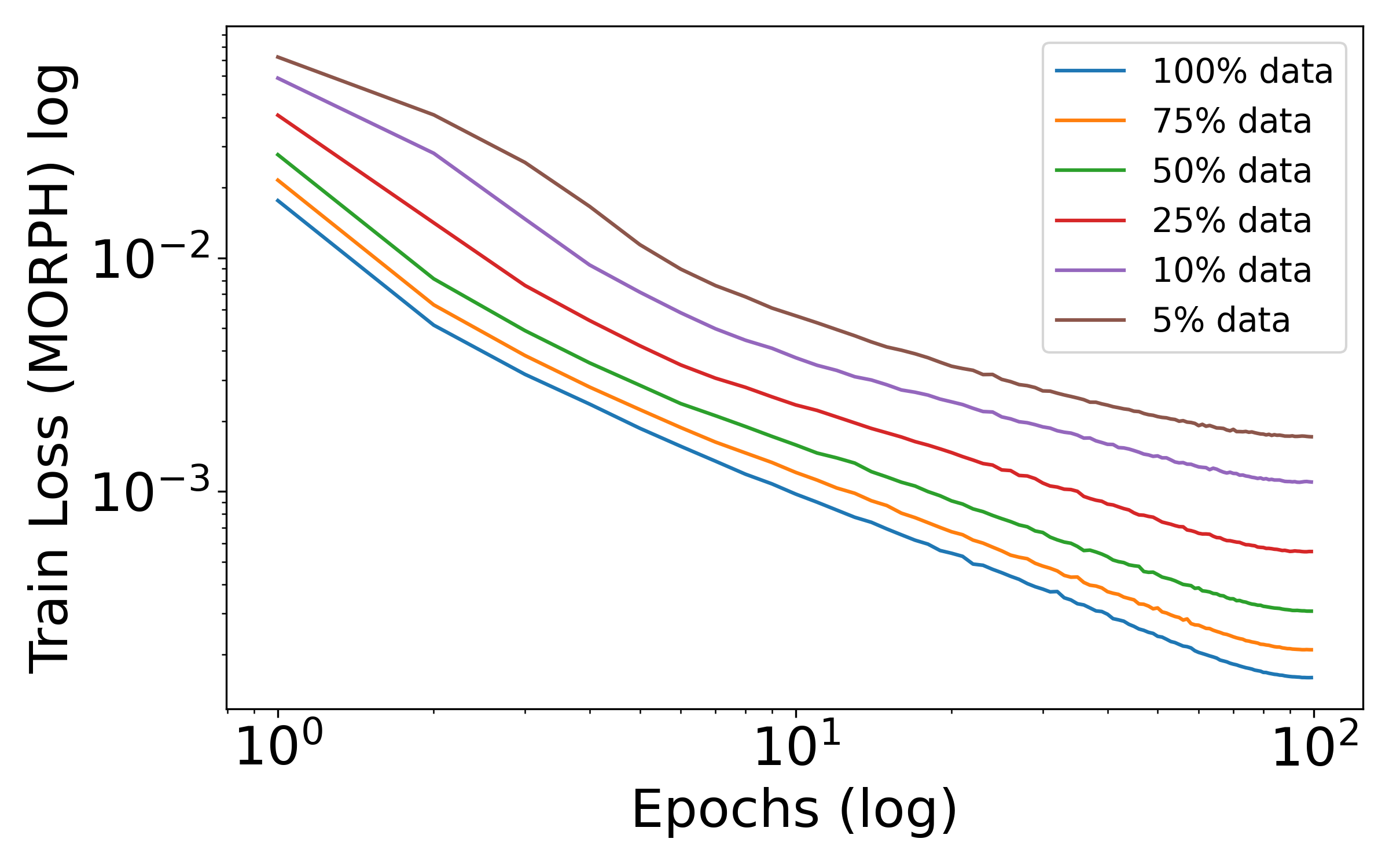}
        \caption{MORPH train loss vs epochs}
        \label{fig:scale_morph_tr}
    \end{subfigure}\hfill
    \begin{subfigure}[b]{0.24\linewidth}
        \centering
        \includegraphics[width=\linewidth]{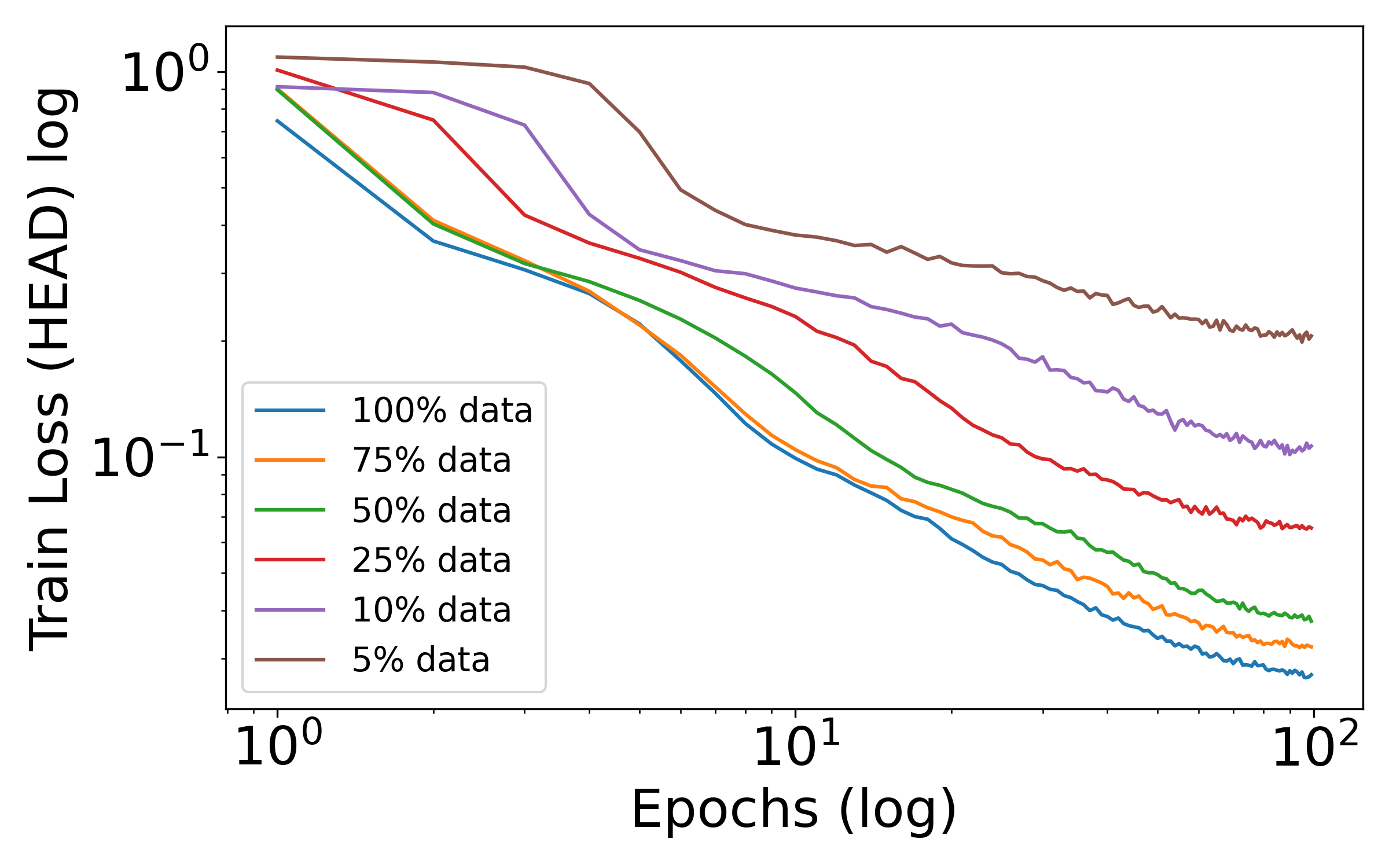}
        \caption{TSH train loss vs epochs}
        \label{fig:scale_head_tr}
    \end{subfigure}\hfill
    \begin{subfigure}[b]{0.24\linewidth}
        \centering
        \includegraphics[width=\linewidth]{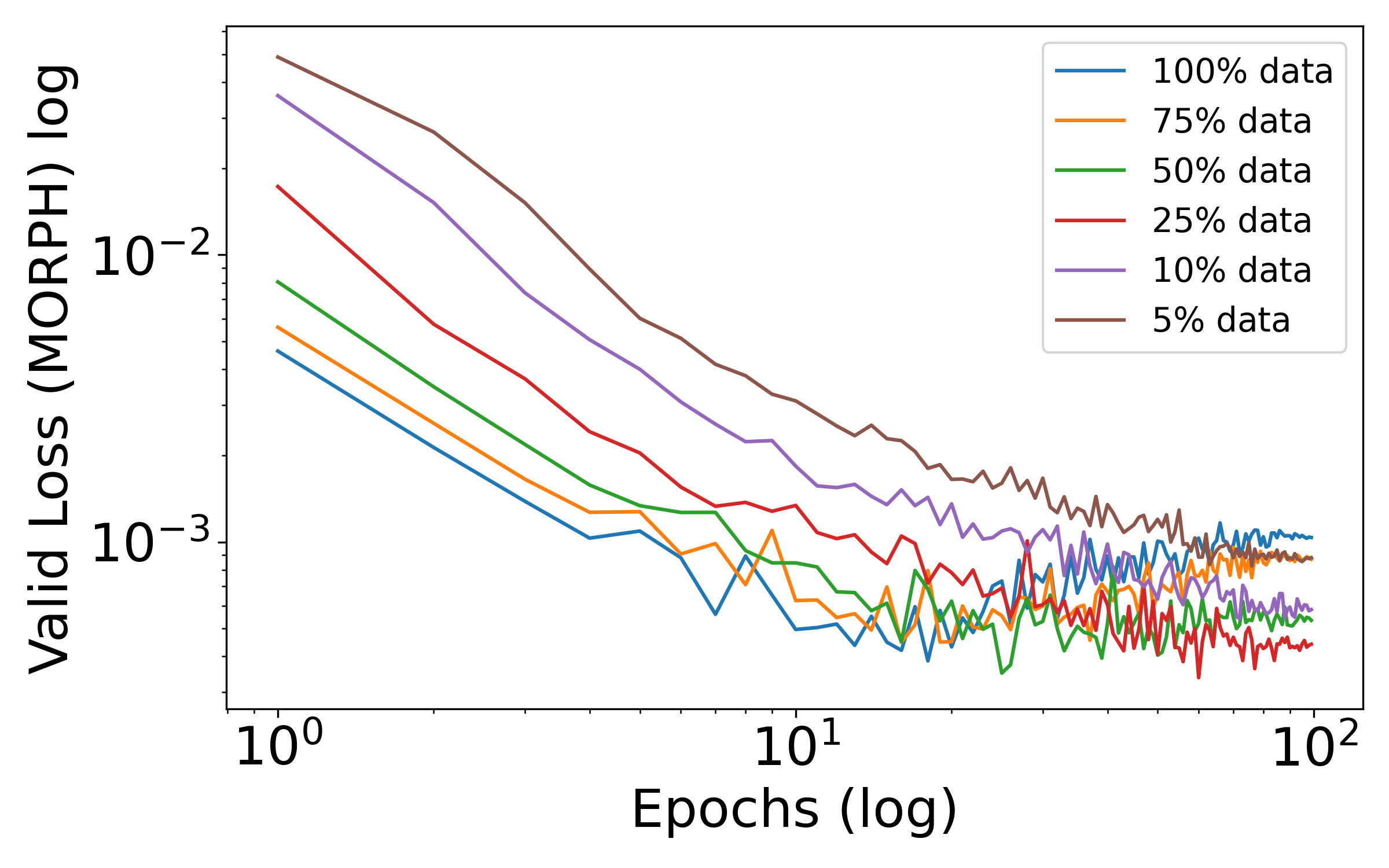}
        \caption{MORPH val loss vs epochs}
        \label{fig:scale_morph_val}
    \end{subfigure}\hfill
    \begin{subfigure}[b]{0.24\linewidth}
        \centering
        \includegraphics[width=\linewidth]{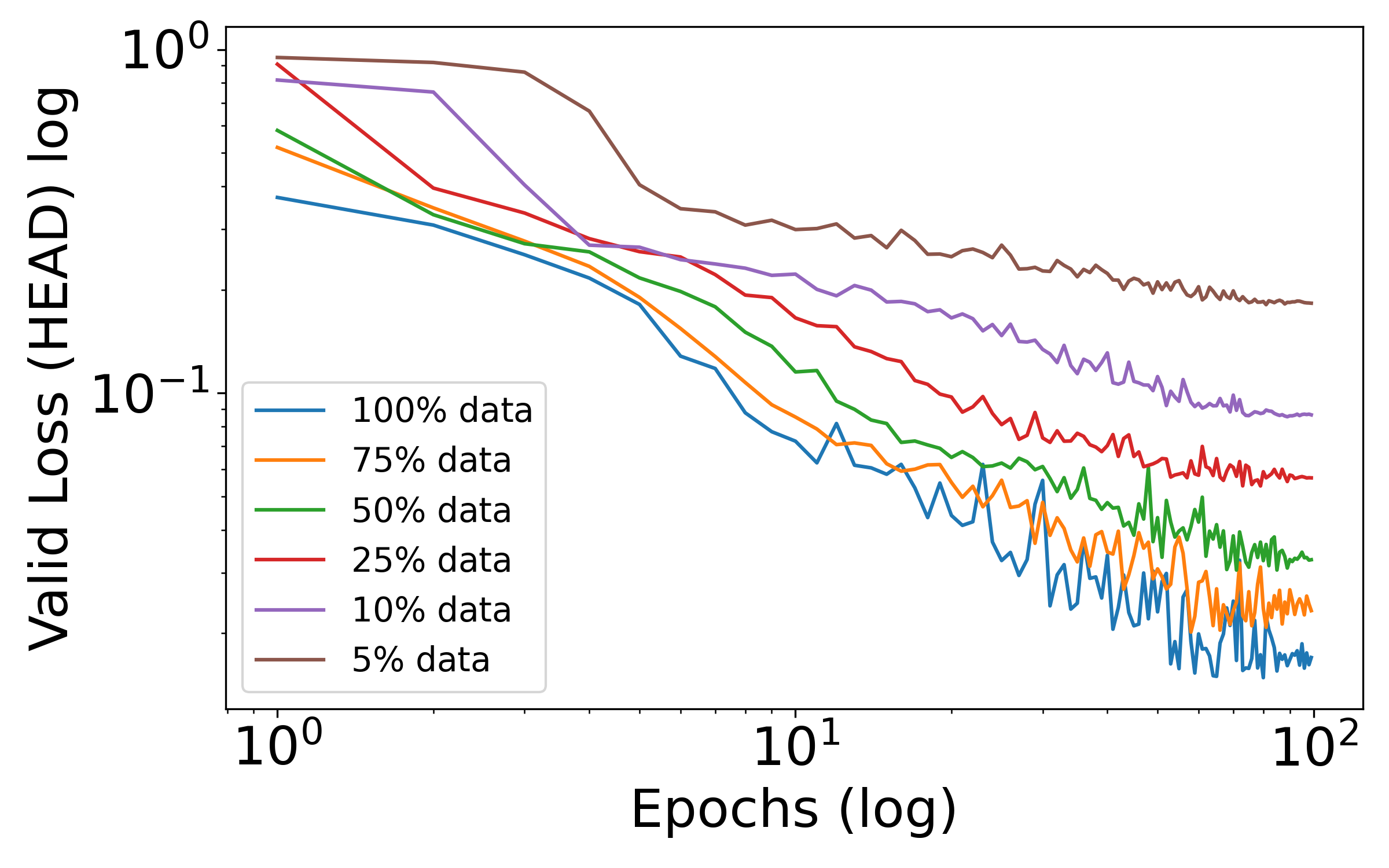}
        \caption{TSH val loss vs epochs}
        \label{fig:scale_head_val}
    \end{subfigure}
    \caption{Data-scaling study showing training and validation loss curves (log-log scale) as the fraction of available training data is varied (5\%, 10\%, 25\%, 50\%, 75\%, and 100 \%). (a) MORPH reconstruction training loss, (b) task-specific head (TSH) parameter-regression training loss, (c) MORPH reconstruction validation loss, and (d) TSH parameter-regression validation loss, all plotted versus epochs (log-scaled axes).}
    \label{fig:scaling}
\end{figure*}

\subsubsection{Hyperspectral image reconstruction}
Fig.~\ref{fig:recon} shows qualitative reconstruction results on four held-out test samples (a–d). For each example, the top row presents the ground-truth four-channel hyperspectral image and the bottom row shows the corresponding reconstruction. MORPH accurately reconstructs hyperspectral test images, including complex multi-lobe structures (e.g., Fig.~\ref{fig:recon_b}). Across the full test set, the reconstruction MSE is $1.2\times 10^{-3}$, indicating that the pretrained backbone learns representations that preserve the essential spectral and spatial structure of the images.

\subsubsection{Parameter estimation}
Fig.~\ref{fig:param_estimation} reports parameter estimation performance on the test set via scatter plots of predicted versus ground-truth values. Due to ill-posedness and limited sensitivity for some inputs, we evaluate three parameters (Param1, Param2, and Param4). The predictions are tightly concentrated around the diagonal, with Param2 exhibiting near-linear agreement across the full range. Quantitatively, Param1 achieves $R^2=0.975$ with $L_2=0.035$, Param2 achieves $R^2=0.995$ with $L_2=0.013$, and Param4 achieves $R^2=0.990$ with $L_2=0.022$. The overall regression MSE on the test set is 0.0235, demonstrating that the fine-tuned MORPH representations support accurate inverse estimation for the identifiable subset of parameters.

\subsubsection{Scaling studies}
We perform data-scaling experiments by varying the fraction of available training data while keeping the same optimization and training hyperparameters. Figure~\ref{fig:scaling} summarizes the training and validation dynamics for six data regimes (5\%, 10\%, 25\%, 50\%, 75\%, and 100\% of the training set), reported on log--log axes. Across both objectives i.e., hyperspectral image reconstruction (MORPH) and parameter regression (TSH) increasing the training set size consistently reduces loss. For MORPH, the reconstruction training loss decreases monotonically with data fraction (Fig.~\ref{fig:scale_morph_tr}), indicating more effective fitting with additional supervision. On the validation set (Fig.~\ref{fig:scale_morph_val}), the largest gains occur when scaling from 5\% to 25\%, after which the curves begin to cluster, suggesting diminishing returns for reconstruction beyond moderate data availability. We also observe mild late-epoch degradation in validation loss for some fractions. In practice, early stopping would mitigate this effect, while the overall scaling trend remains clear. In contrast, the task-specific head exhibits sustained improvements with more data: both the regression training loss (Fig.~\ref{fig:scale_head_tr}) and validation loss (Fig.~\ref{fig:scale_head_val}) continue to decrease up to 75--100\% of the data.  

Overall, these results confirm that additional data yields more stable training dynamics and improvement in loss for both reconstruction and parameter estimation. These trends suggest further gains are likely as larger ICF datasets become available.

\begin{figure}
	\centering
	\includegraphics[width=0.5\textwidth,trim=0cm 0cm 0cm 0cm, clip]{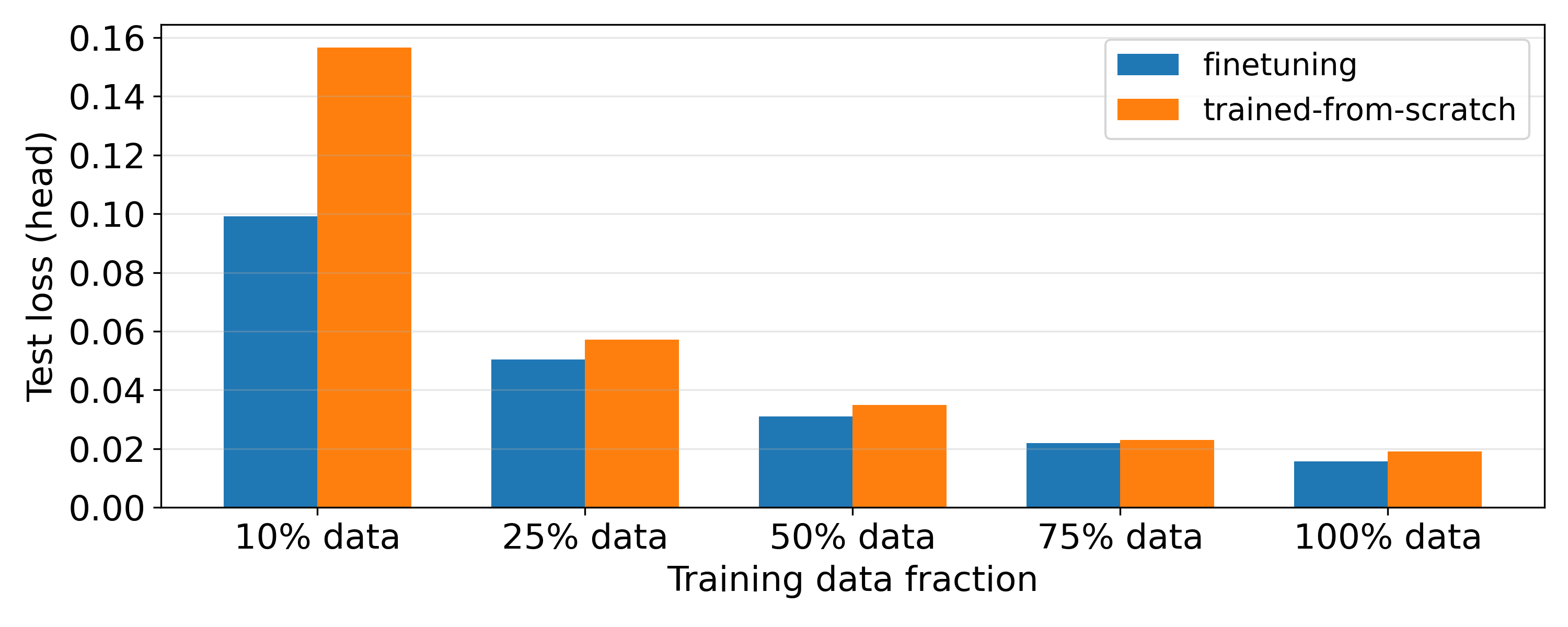}
	\caption{fine-tuned model vs trained-from-scratch: Test-set parameter-regression loss of the TSH versus training-data fraction, comparing MORPH finetuning from pretrained weights to training the same architecture from scratch. Finetuning achieves lower loss, with the largest gains in the low-data regime (e.g., 10\%), and the gap narrows as more training data becomes available.}
	\label{fig:compare}
	\vspace{2mm}
\end{figure}

\subsection{Comparisons}
We compare MORPH fine-tuning with a matched training-from-scratch baseline that uses the identical architecture and hyperparameters, with the only difference being whether weights are initialized from pretrained checkpoints. This controlled comparison isolates the effect of pretraining and quantifies its benefit in a data- and compute-limited regime. Fig.~\ref{fig:compare} compares the test-set parameter-regression loss of the task-specific head (TSH) when (i) finetuning MORPH from pretrained weights versus (ii) training the identical architecture from scratch, across multiple training-data fractions. Finetuning yields lower test loss in the most data-limited regime, with a pronounced improvement at 10\% of the data, highlighting the sample-efficiency benefit of pretraining. As the training fraction increases (25\%–100\%), the gap progressively narrows, although finetuning remains consistently better, suggesting that pretraining provides the greatest advantage when supervision is scarce and its relative benefit diminishes as more labeled data becomes available. 

Overall, these results show that adding training data improves both reconstruction and parameter estimation, while also stabilizing the learning dynamics. This scaling behavior suggests that further gains are likely as larger ICF datasets become available.

\section*{Conclusions}
We investigated PDE foundation-model transfer for an inverse ICF parameter-estimation task using the JAG benchmark, where multi-modal diagnostics (hyperspectral images and scalar observables) are used to infer simulator input parameters. A sensitivity analysis indicated that two parameters are weakly identifiable under the available observables, making the full inversion ill-posed and motivating a reduced task focused on three parameters. Finetuning MORPH-S with a lightweight task-specific head achieves accurate hyperspectral reconstruction (test MSE $1.2\times10^{-3}$) and strong parameter-estimation performance (up to $R^2=0.995$ on the best-predicted parameter). Data-scaling and baseline comparisons show that pretraining improves sample efficiency, yielding lower test loss than training from scratch, with the largest gains in the low-data regime. These results suggest PDE foundation models are a promising building block for data-limited inverse problems in ICF, and motivate future work on improving sensitivity via a larger training set, richer diagnostics, stronger priors, and more robust inverse objectives.

\bibliographystyle{IEEEtranN}
\bibliography{bibfile}

@article{rautela2025morph,
  title={MORPH: PDE Foundation Models with Arbitrary Data Modality},
  author={Rautela, Mahindra Singh and Most, Alexander and Mansingh, Siddharth and Love, Bradley C and Biswas, Ayan and Oyen, Diane and Lawrence, Earl},
  journal={arXiv preprint arXiv:2509.21670},
  year={2025}
}

@article{bodnar2025foundation,
  title={A foundation model for the Earth system},
  author={Bodnar, Cristian and Bruinsma, Wessel P and Lucic, Ana and Stanley, Megan and Allen, Anna and Brandstetter, Johannes and Garvan, Patrick and Riechert, Maik and Weyn, Jonathan A and Dong, Haiyu and others},
  journal={Nature},
  pages={1--8},
  year={2025},
  publisher={Nature Publishing Group UK London}
}

@article{brown2020language,
  title={Language models are few-shot learners},
  author={Brown, Tom and Mann, Benjamin and Ryder, Nick and Subbiah, Melanie and Kaplan, Jared D and Dhariwal, Prafulla and Neelakantan, Arvind and Shyam, Pranav and Sastry, Girish and Askell, Amanda and others},
  journal={Advances in neural information processing systems},
  volume={33},
  pages={1877--1901},
  year={2020}
}

@article{marcato2025foundation,
  title={A Foundation Model for Material Fracture Prediction},
  author={Marcato, Agnese and Pachalieva, Aleksandra and Hill, Ryley G and Gao, Kai and Wang, Xiaoyu and Rougier, Esteban and Lei, Zhou and Agrawal, Vinamra and Chua, Janel and Kang, Qinjun and others},
  journal={arXiv preprint arXiv:2507.23077},
  year={2025}
}

@article{oquab2023dinov2,
  title={Dinov2: Learning robust visual features without supervision},
  author={Oquab, Maxime and Darcet, Timoth{\'e}e and Moutakanni, Th{\'e}o and Vo, Huy and Szafraniec, Marc and Khalidov, Vasil and Fernandez, Pierre and Haziza, Daniel and Massa, Francisco and El-Nouby, Alaaeldin and others},
  journal={arXiv preprint arXiv:2304.07193},
  year={2023}
}

@article{wadell2025foundation,
  title={Foundation models for discovery and exploration in chemical space},
  author={Wadell, Alexius and Bhutani, Anoushka and Azumah, Victor and Ellis-Mohr, Austin R and Kelly, Celia and Zhao, Hancheng and Nayak, Anuj K and Hegazy, Kareem and Brace, Alexander and Lin, Hongyi and others},
  journal={arXiv preprint arXiv:2510.18900},
  year={2025}
}

@article{anirudh2020improved,
  title={Improved surrogates in inertial confinement fusion with manifold and cycle consistencies},
  author={Anirudh, Rushil and Thiagarajan, Jayaraman J and Bremer, Peer-Timo and Spears, Brian K},
  journal={Proceedings of the National Academy of Sciences},
  volume={117},
  number={18},
  pages={9741--9746},
  year={2020},
  publisher={National Academy of Sciences}
}

@article{anirudh2019exploring,
  title={Exploring generative physics models with scientific priors in inertial confinement fusion},
  author={Anirudh, Rushil and Thiagarajan, Jayaraman J and Liu, Shusen and Bremer, Peer-Timo and Spears, Brian K},
  journal={arXiv preprint arXiv:1910.01666},
  year={2019}
}

@inproceedings{jacobs2019parallelizing,
  title={Parallelizing training of deep generative models on massive scientific datasets},
  author={Jacobs, Sam Ade and Van Essen, Brian and Hysom, David and Yeom, Jae-Seung and Moon, Tim and Anirudh, Rushil and Thiagaranjan, Jayaraman J and Liu, Shusen and Bremer, Peer-Timo and Gaffney, Jim and others},
  booktitle={2019 IEEE International Conference on Cluster Computing (CLUSTER)},
  pages={1--10},
  year={2019},
  organization={IEEE}
}

@article{shukla2025design,
  title={On the design and evaluation of generative models in high energy density physics},
  author={Shukla, Ankita and Mubarka, Yamen and Anirudh, Rushil and Kur, Eugene and Mariscal, Derek and Djordjevic, Blagoje and Kustowski, Bogdan and Swanson, Kelly and Spears, Brian and Bremer, Peer-Timo and others},
  journal={Communications Physics},
  volume={8},
  number={1},
  pages={14},
  year={2025},
  publisher={Nature Publishing Group UK London}
}

@article{rautela2022inverse,
  title={Inverse characterization of composites using guided waves and convolutional neural networks with dual-branch feature fusion},
  author={Rautela, Mahindra and Huber, Armin and Senthilnath, J and Gopalakrishnan, S},
  journal={Mechanics of Advanced Materials and Structures},
  volume={29},
  number={27},
  pages={6595--6611},
  year={2022},
  publisher={Taylor \& Francis}
}

@book{vogel2002computational,
  title={Computational methods for inverse problems},
  author={Vogel, Curtis R},
  year={2002},
  publisher={SIAM}
}

@article{mansingh2025towards,
  title={Towards Reasoning for PDE Foundation Models: A Reward-Model-Driven Inference-Time-Scaling Algorithm},
  author={Mansingh, Siddharth and Amarel, James and Arnab, Ragib and Mohan, Arvind and Singh, Kamaljeet and Kunde, Gerd J and Hengartner, Nicolas and Migliori, Benjamin and Casleton, Emily and Debardeleben, Nathan A and others},
  journal={arXiv preprint arXiv:2509.02846},
  year={2025}
}

@article{balasubramaniam1998inversion,
  title={Inversion of the ply lay-up sequence for multi-layered fiber reinforced composite plates using genetic algorithm},
  author={Balasubramaniam, Krishnan},
  journal={Nondestructive Testing and Evaluation},
  volume={15},
  number={5},
  pages={311--331},
  year={1998},
  publisher={Taylor \& Francis}
}

@article{om2001ridge,
  title={Ridge regression and inverse problems},
  author={Björkström, Andres},
  journal={Stockholm University, Department of Mathematics},
  year={2001}
}

@inproceedings{bahukutumbi2025fine,
  title={Fine-tuning Large Language Models for Inertial Confinement Fusion--a collaboration with NVIDIA},
  author={Bahukutumbi, Radha and Meyerhofer, David D and Debardeleben, Nathan and Lang, Michael and Jhunjhunwala, Aastha and Vem, Avinash and Luu, Douglas and Gurumurthy, Prakash and Halverson, Scott and Gupta, Geetika and others},
  booktitle={DPP 2025},
  year={2025},
  organization={APS}
}

@article{grosskopf2025ursa,
  title={URSA: The Universal Research and Scientific Agent},
  author={Grosskopf, Michael and Bent, Russell and Somasundaram, Rahul and Michaud, Isaac and Lui, Arthur and Debardeleben, Nathan and Lawrence, Earl},
  journal={arXiv preprint arXiv:2506.22653},
  year={2025}
}

@inproceedings{gutierrez2025lpg,
  title={IM-LPG: Inverse Modeling Approach to Laser Pulse Shape Generation in Inertial Confinement Fusion},
  author={Gutierrez, Ricardo Luna and Gundecha, Vineet and Ejaz, Rahman and Gopalaswamy, Varchas and Betti, Riccardo and Ghorbanpour, Sahand and Lees, Aarne and Sarkar, Soumyendu},
  booktitle={NeurIPS 2025 AI for Science Workshop},
  year={2025}
}

@article{serino2025physics,
  title={Physics consistent machine learning framework for inverse modeling with applications to ICF capsule implosions},
  author={Serino, Daniel A and Bell, Evan and Klasky, Marc and Southworth, Ben S and Nadiga, Balasubramanya and Wilcox, Trevor and Korobkin, Oleg},
  journal={Scientific Reports},
  volume={15},
  number={1},
  pages={25915},
  year={2025},
  publisher={Nature Publishing Group UK London}
}

@article{gaffney2024data,
  title={Data-driven prediction of scaling and ignition of inertial confinement fusion experiments},
  author={Gaffney, Jim A and Humbird, Kelli and Kritcher, Andrea and Kruse, Michael and Kur, Eugene and Kustowski, Bogdan and Nora, Ryan and Spears, Brian},
  journal={Physics of Plasmas},
  volume={31},
  number={9},
  year={2024},
  publisher={AIP Publishing}
}

@article{humbird2022transfer,
  title={Transfer learning driven design optimization for inertial confinement fusion},
  author={Humbird, Kelli D and Peterson, Jayson Luc},
  journal={Physics of Plasmas},
  volume={29},
  number={10},
  year={2022},
  publisher={AIP Publishing}
}

@article{wang2024multifidelity,
  title={A multifidelity Bayesian optimization method for inertial confinement fusion design},
  author={Wang, Jingyi and Chiang, N and Gillette, Andrew and Peterson, J Luc},
  journal={Physics of Plasmas},
  volume={31},
  number={3},
  year={2024},
  publisher={AIP Publishing}
}

@misc{JAG_LLNL,
	title = {The JAG inertial confinement fusion simulation dataset for multi-modal scientific deep learning.},
	author = {Gaffney, Jim A. and Anirudh, Rushil and Bremer, Peer-Timo and Hammer, Jim and Hysom, David and Jacobs, Sam A. and Peterson, J. Luc and Robinson, Peter and Spears, 	Brian K. and Springer, Paul T. and Thiagarajan, Jayaraman J. and Van Essen, Brian and Yeom, Jae-Seung},
	url = {https://library.ucsd.edu/dc/object/bb5534097t},
	howpublished = {In Lawrence Livermore National Laboratory (LLNL) Open Data Initiative. UC San Diego Library Digital Collections.},
	doi = {https://doi.org/10.6075/J0RV0M27},
	year = {2020},
	month = {3}
}

@inproceedings{gaffney2014thermodynamic,
  title={Thermodynamic modeling of uncertainties in NIF ICF implosions due to underlying microphysics models},
  author={Gaffney, Jim and Springer, Paul and Collins, Gilbert},
  booktitle={APS Division of Plasma Physics Meeting Abstracts},
  volume={2014},
  pages={PO5--011},
  year={2014}
}

@article{RAISSI2019686,
title = {Physics-informed neural networks: A deep learning framework for solving forward and inverse problems involving nonlinear partial differential equations},
journal = {Journal of Computational Physics},
volume = {378},
pages = {686-707},
year = {2019},
issn = {0021-9991},
author = {M. Raissi and P. Perdikaris and G.E. Karniadakis},
}

@article{li2024physics,
  title={Physics-informed neural operator for learning partial differential equations},
  author={Li, Zongyi and Zheng, Hongkai and Kovachki, Nikola and Jin, David and Chen, Haoxuan and Liu, Burigede and Azizzadenesheli, Kamyar and Anandkumar, Anima},
  journal={ACM/JMS Journal of Data Science},
  volume={1},
  number={3},
  pages={1--27},
  year={2024},
  publisher={ACM New York, NY}
}

@article{oommen2022learning,
  title={Learning two-phase microstructure evolution using neural operators and autoencoder architectures},
  author={Oommen, Vivek and Shukla, Khemraj and Goswami, Somdatta and Dingreville, R{\'e}mi and Karniadakis, George Em},
  journal={npj Computational Materials},
  volume={8},
  number={1},
  pages={190},
  year={2022},
  publisher={Nature Publishing Group UK London}
}

@article{maulik2021reduced,
  title={Reduced-order modeling of advection-dominated systems with recurrent neural networks and convolutional autoencoders},
  author={Maulik, Romit and Lusch, Bethany and Balaprakash, Prasanna},
  journal={Physics of Fluids},
  volume={33},
  number={3},
  year={2021},
  publisher={AIP Publishing}
}

@article{kontolati2024learning,
  title={Learning nonlinear operators in latent spaces for real-time predictions of complex dynamics in physical systems},
  author={Kontolati, Katiana and Goswami, Somdatta and Em Karniadakis, George and Shields, Michael D},
  journal={Nature Communications},
  volume={15},
  number={1},
  pages={5101},
  year={2024},
  publisher={Nature Publishing Group UK London}
}

@article{khodkar2021data,
  title={A data-driven, physics-informed framework for forecasting the spatiotemporal evolution of chaotic dynamics with nonlinearities modeled as exogenous forcings},
  author={Khodkar, MA and Hassanzadeh, Pedram},
  journal={Journal of Computational Physics},
  volume={440},
  pages={110412},
  year={2021},
  publisher={Elsevier}
}

@article{chen2018neural,
  title={Neural ordinary differential equations},
  author={Chen, Ricky TQ and Rubanova, Yulia and Bettencourt, Jesse and Duvenaud, David K},
  journal={Advances in neural information processing systems},
  volume={31},
  year={2018}
}

@article{li2020fourier,
  title={Fourier neural operator for parametric partial differential equations},
  author={Li, Zongyi and Kovachki, Nikola and Azizzadenesheli, Kamyar and Liu, Burigede and Bhattacharya, Kaushik and Stuart, Andrew and Anandkumar, Anima},
  journal={arXiv preprint arXiv:2010.08895},
  year={2020}
}

@article{rautela2024conditional,
  title={A conditional latent autoregressive recurrent model for generation and forecasting of beam dynamics in particle accelerators},
  author={Rautela, Mahindra and Williams, Alan and Scheinker, Alexander},
  journal={Scientific Reports},
  volume={14},
  number={1},
  pages={18157},
  year={2024},
  publisher={Nature Publishing Group UK London}
}

@article{rautela2025time,
  title={Time-inversion of spatiotemporal beam dynamics using uncertainty-aware latent evolution reversal},
  author={Rautela, Mahindra and Williams, Alan and Scheinker, Alexander},
  journal={Physical Review E},
  volume={111},
  number={2},
  pages={025307},
  year={2025},
  publisher={APS}
}

@article{goswami2022deep,
  title={Deep transfer operator learning for partial differential equations under conditional shift},
  author={Goswami, Somdatta and Kontolati, Katiana and Shields, Michael D and Karniadakis, George Em},
  journal={Nature Machine Intelligence},
  volume={4},
  number={12},
  pages={1155--1164},
  year={2022},
  publisher={Nature Publishing Group UK London}
}

@article{tripura2023wavelet,
  title={Wavelet neural operator for solving parametric partial differential equations in computational mechanics problems},
  author={Tripura, Tapas and Chakraborty, Souvik},
  journal={Computer Methods in Applied Mechanics and Engineering},
  volume={404},
  pages={115783},
  year={2023},
  publisher={Elsevier}
}

@article{cao2024laplace,
  title={Laplace neural operator for solving differential equations},
  author={Cao, Qianying and Goswami, Somdatta and Karniadakis, George Em},
  journal={Nature Machine Intelligence},
  volume={6},
  number={6},
  pages={631--640},
  year={2024},
  publisher={Nature Publishing Group UK London}
}

@article{takamoto2022pdebench,
  title={Pdebench: An extensive benchmark for scientific machine learning},
  author={Takamoto, Makoto and Praditia, Timothy and Leiteritz, Raphael and MacKinlay, Daniel and Alesiani, Francesco and Pfl{\"u}ger, Dirk and Niepert, Mathias},
  journal={Advances in Neural Information Processing Systems},
  volume={35},
  pages={1596--1611},
  year={2022}
}

@article{ohana2024well,
  title={The well: a large-scale collection of diverse physics simulations for machine learning},
  author={Ohana, Ruben and McCabe, Michael and Meyer, Lucas and Morel, Rudy and Agocs, Fruzsina and Beneitez, Miguel and Berger, Marsha and Burkhart, Blakesly and Dalziel, Stuart and Fielding, Drummond and others},
  journal={Advances in Neural Information Processing Systems},
  volume={37},
  pages={44989--45037},
  year={2024}
}

@article{hao2024dpot,
  title={Dpot: Auto-regressive denoising operator transformer for large-scale pde pre-training},
  author={Hao, Zhongkai and Su, Chang and Liu, Songming and Berner, Julius and Ying, Chengyang and Su, Hang and Anandkumar, Anima and Song, Jian and Zhu, Jun},
  journal={arXiv preprint arXiv:2403.03542},
  year={2024}
}

@article{herde2024poseidon,
  title={Poseidon: Efficient foundation models for pdes},
  author={Herde, Maximilian and Raonic, Bogdan and Rohner, Tobias and K{\"a}ppeli, Roger and Molinaro, Roberto and de B{\'e}zenac, Emmanuel and Mishra, Siddhartha},
  journal={Advances in Neural Information Processing Systems},
  volume={37},
  pages={72525--72624},
  year={2024}
}

@article{subramanian2023towards,
  title={Towards foundation models for scientific machine learning: Characterizing scaling and transfer behavior},
  author={Subramanian, Shashank and Harrington, Peter and Keutzer, Kurt and Bhimji, Wahid and Morozov, Dmitriy and Mahoney, Michael W and Gholami, Amir},
  journal={Advances in Neural Information Processing Systems},
  volume={36},
  pages={71242--71262},
  year={2023}
}

\end{document}